\documentclass[twoside,12pt,a4paper]{report}
\usepackage{epsf}
\usepackage{url}
\usepackage{graphics, graphicx}
\usepackage{latexsym}
\usepackage[margin=10pt,font=small,labelfont=bf]{caption}
\usepackage[utf8]{inputenc}
\usepackage[toc,page]{appendix}
\usepackage[doublespacing]{setspace}
\usepackage{listings}

\usepackage{color}
\usepackage{amsthm}
\usepackage{xcolor}

\usepackage{verbatim}

\usepackage{tikz}

\definecolor{dkgreen}{rgb}{0,0.6,0}
\definecolor{gray}{rgb}{0.5,0.5,0.5}
\definecolor{mauve}{rgb}{0.58,0,0.82}
\definecolor{brown}{rgb}{0.80,0.72,0.44}
\definecolor{darkbrown}{rgb}{0.75,0.67,0.39}
\definecolor{red}{rgb}{1,0.40,0.42}

\lstdefinestyle{myHaskellStyle}{
  frame=tb,
  language=haskell,
  aboveskip=3mm,
  belowskip=3mm,
  showstringspaces=false,
  columns=flexible,
  basicstyle={\small\ttfamily},
  numbers=none,
  numberstyle=\color{darkbrown},
  keywordstyle=\color{blue},
  commentstyle=\color{dkgreen},
  stringstyle=\color{mauve},
  frame=single,
  breaklines=true,
  breakatwhitespace=true,
  tabsize=3,
  keywords=[1]{case,class,data,deriving,do,else,if,import,in,infixl,infixr,instance,let,module,of,primitive,then,type,where,family,newtype},
  keywordstyle=[1]\color{blue},
  morekeywords=[2]{->,|,=>,::,\,*,<-},
  otherkeywords={->,|,=>,::,\,*,<-},
  keywordstyle=[2]\color{mauve},
  morekeywords=[3]{Move,Player,Board,R,Int,Matrix,Bool,Tree},
  keywordstyle=[3]\color{brown},
  morekeywords=[4]{N, X, O,False,True,Node,Leaf},
  keywordstyle=[4]\color{red},
}

\lstdefinestyle{myHaskellStyleInline}{
  frame=tb,
  language=haskell,
  aboveskip=0mm,
  belowskip=0mm,
  showstringspaces=false,
  columns=flexible,
  basicstyle={\small\ttfamily},
  numbers=none,
  numberstyle=\color{darkbrown},
  keywordstyle=\color{blue},
  commentstyle=\color{dkgreen},
  stringstyle=\color{mauve},
  frame=single,
  breaklines=true,
  breakatwhitespace=true,
  tabsize=3,
  keywords=[1]{case,class,data,deriving,do,else,if,import,in,infixl,infixr,instance,let,module,of,primitive,then,type,where,family,newtype},
  keywordstyle=[1]\color{blue},
  morekeywords=[2]{->,|,=>,::,\,*,<-},
  otherkeywords={->,|,=>,::,\,*,<-},
  keywordstyle=[2]\color{mauve},
  morekeywords=[3]{Move,Player,Board,R,Int,Matrix,Bool,Tree},
  keywordstyle=[3]\color{brown},
  morekeywords=[4]{N, X, O,False,True,Node,Leaf},
  keywordstyle=[4]\color{red},
}

\usepackage{graphicx}
\tikzset{
  treenode/.style = {align=center, inner sep=0pt, text centered,
    font=\sffamily},
  arn_n/.style = {treenode, circle, black, font=\sffamily\bfseries, draw=black,
    fill=white, text width=1.5em},% arbre rouge noir, noeud noir
  arn_r/.style = {treenode, circle, red, draw=red, 
    text width=1.5em, very thick},
}
\usetikzlibrary{arrows,automata}

\begin{document}
\singlespacing
%%%%%%%%%%%%%%%%%%%%%%%%%%%%%%%%%%%%%%%%%%%%%%%%%%%%%%%%%%%%%%%%%%%%%%%%%%%%
%%% hier steht die neue Titelseite 
%%%%%%%%%%%%%%%%%%%%%%%%%%%%%%%%%%%%%%%%%%%%%%%%%%%%%%%%%%%%%%%%%%%%%%%%%%%%
 
\begin{titlepage}
\begin{tabular}{p{8cm} | p{8cm}}

 	\vspace{1.5cm}\textbf{School of Electronic \newline Engineering and \newline Computer Science} & \vspace{1.5cm} MSc Software Engineering \\
  & \vspace{3cm} Finding optimal strategies in \newline sequential games with the novel \newline selection monad   \cr \vspace{7cm} \includegraphics[width = .5\textwidth]{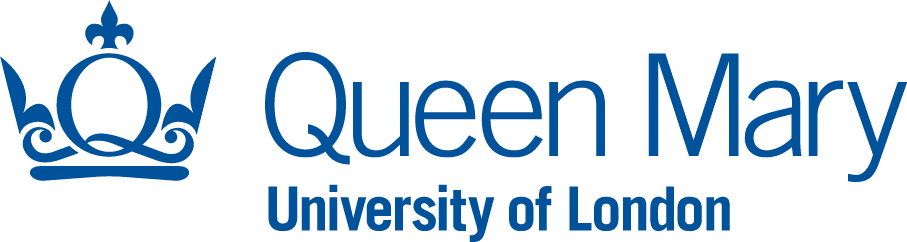} & \vspace{9cm}  Johannes Niklas Hartmann  \newline August 22, 2018 \\ 
  \vspace{3cm}

\end{tabular}

\vfill

\normalsize

\newpage
\end{titlepage}

%%%%%%%%%%%%%%%%%%%%%%%%%%%%%%%%%%%%%%%%%%%%%%%%%%%%%%%%%%%%%%%%%%%%%%%%%%%%
%%% Titelr"uckseite: Bibliographische Angaben
%%%%%%%%%%%%%%%%%%%%%%%%%%%%%%%%%%%%%%%%%%%%%%%%%%%%%%%%%%%%%%%%%%%%%%%%%%%%

\thispagestyle{empty}
\vspace*{\fill}
\begin{minipage}{11.2cm}
\textbf{Hartmann, Johannes Niklas:}\\
\emph{Finding optimal strategies in sequential \newline games with the novel selection monad }\\ Master Thesis Software Engineering\\
Queen Mary University London \\
\end{minipage}
\newpage

%%%%%%%%%%%%%%%%%%%%%%%%%%%%%%%%%%%%%%%%%%%%%%%%%%%%%%%%%%%%%%%%%%%%%%%%%%%%

\pagenumbering{roman}
\setcounter{page}{1}

%%%%%%%%%%%%%%%%%%%%%%%%%%%%%%%%%%%%%%%%%%%%%%%%%%%%%%%%%%%%%%%%%%%%%%%%%%%%
%%% Seite I: Zusammenfassug, Danksagung
%%%%%%%%%%%%%%%%%%%%%%%%%%%%%%%%%%%%%%%%%%%%%%%%%%%%%%%%%%%%%%%%%%%%%%%%%%%%

\doublespacing
\section*{Abstract}
The recently discovered monad, \verb+Tx = Selection (x -> r) -> r+, provides an elegant way to find optimal strategies in sequential games. During this thesis, a library was developed which provides a set of useful functions using the selection monad to compute optimal games and AIs for sequential games. In order to explore the selection monads ability to support this AI implementations, three example case studies where developed using Haskell: The two player game Connect Four, a Sudoku solver and a simplified version of Chess. These case studies show how to elegantly implement a game AI. Furthermore, a performance analysis of this case studies was done, identifying the major points where performance can be increased.

\newpage

%%%%%%%%%%%%%%%%%%%%%%%%%%%%%%%%%%%%%%%%%%%%%%%%%%%%%%%%%%%%%%%%%%%%%%%%%%%%%
%%% Table of Contents
%%%%%%%%%%%%%%%%%%%%%%%%%%%%%%%%%%%%%%%%%%%%%%%%%%%%%%%%%%%%%%%%%%%%%%%%%%%%%
\singlespacing
\renewcommand{\baselinestretch}{1.3}
\small\normalsize

\tableofcontents

\renewcommand{\baselinestretch}{1}
\small\normalsize

%%%%%%%%%%%%%%%%%%%%%%%%%%%%%%%%%%%%%%%%%%%%%%%%%%%%%%%%%%%%%%%%%%%%%%%%%%%%%
%%% List of Figures
%%%%%%%%%%%%%%%%%%%%%%%%%%%%%%%%%%%%%%%%%%%%%%%%%%%%%%%%%%%%%%%%%%%%%%%%%%%%%

\renewcommand{\baselinestretch}{1.3}
\small\normalsize

\addcontentsline{toc}{chapter}{List of Figures}
\listoffigures

\renewcommand{\baselinestretch}{1}
\small\normalsize

%%%%%%%%%%%%%%%%%%%%%%%%%%%%%%%%%%%%%%%%%%%%%%%%%%%%%
%%% Der Haupttext, ab hier mit arabischer Numerierung
%%% Mit \input{dateiname} werden die Datei `dateiname' eingebunden
%%%%%%%%%%%%%%%%%%%%%%%%%%%%%%%%%%%%%%%%%%%%%%%%%%%%%%%%%%%%%%%%%%%%%%%%%%%%%

\pagenumbering{arabic}
\setcounter{page}{1}

%%%%%%%%%%%%%%%%%%%%%%%%%%%%%%%%%%%%%%%%%%%%%%%%%%%%%%%%%%%%%%%%%%%%
% Einleitung
%%%%%%%%%%%%%%%%%%%%%%%%%%%%%%%%%%%%%%%%%%%%%%%%%%%%%%%%%%%%%%%%%%%%
% 3 Pages

\doublespacing
\chapter{Introduction}\label{Introduction}
The recently discovered monad, \verb+Tx = Selection (x -> r) -> r+, can be used to elegantly find optimal strategies in sequential games. It can also be used to implement a computational version of the Tychonoff Theorem and to realize the Double-Negation Shift \cite{SequtentialGames}. However, the aim of this master thesis is to explore the monad's ability to find optimal strategies in sequential games by implementing a collection of example games. These games can be seen as case studies that use the \textit{selection monad} and are implemented in the programming language Haskell. Furthermore, common functionalities in these case studies were examined and summarised in a library.
%TODO Noch etwas allgemeiner und paulo erwähnen
\section{Goal of the thesis}
The goal of this thesis is to explore the ability of the \textit{selection monad} to support elegant implementations of AIs playing sequential games. The approach of this AI is to explore all possible moves for each player and to find a sequence of moves that lead to the optimal outcome. This project aims to explore the monad's ability to do so, by implementing a collection of case studies. The feasibility of this monad to implement these case studies will be investigated in terms of performance of the implementation and suitability of the \textit{selection monad} for this case study. Furthermore, with the experience of implementing these case studies, a library was built. This library will provide future programmers an elegant and modular way of implementing AIs for sequential games.\\
Mainly, this thesis consists of the following parts:
\begin{itemize}
	\item Background research
	\item Library implementation
	\item Implementation of three case studies
	\item Performance analysis of the case studies
	\item Discussion of performance and suitability of the selection monad
\end{itemize}
\subsection{Case studies}
The case study games for this project are Connect Four, Sudoku and a simplified version of Chess. These games are all sequential games, which means a player has unlimited time to think about his move and eventually ends his turn by doing the move. These case studies are chosen to investigate different challenges that need to be solved in the games implementation. Connect four is a classic example of a two-player game with three possible outcomes for each player (Win, Draw, Loose). Sudoku, on the other hand, is a one-player game with only two possible outcomes (Valid and Invalid board). The simplified version of chess concentrates on solving chess end-games with limited pieces (only Queen, King, Rook and Bishop). Therefore, the challenge is to show that even a sophisticated game like chess can be implemented to some extent with the help of the \textit{selection monad}. The case studies will be introduced in more detail in Section \ref{ExampleGames}.
\subsection{Library}
As a result of the case studies, a library was built that contains all \textit{selection monad} specific implementations, as well as some helper functions that can be used to build the AI for a sequential game. The goal of this library is to summarise all common functionalities of the case studies, in order to avoid code duplicates and also enable a modular programming approach. The library contains the definition of the \textit{selection monad} with the related functions, a set of minimum and maximum functions used to describe the behaviour of the AI, and functions that compute the next move of the AI, as well as optimal plays and the corresponding optimal outcome. The design of the library will be explained in detail in Section \ref{Library}.
\subsection{Performance}
In order to compute an optimal play for a sequential game, a computer program needs to explore all possible moves. Because this can be a lot of moves (for example connect four has 4,531,985,219,092 different possible games \cite{ConnectFour}), it is important to take performance optimisations into account when implementing these example games.  
Some optimisations can be implemented in the min or max functions of the library. However, it is also important to implement efficient functions that calculate the outcome of a game, as well as calculating all possible moves. Due to the high number of possible moves, it is often impossible to explore all possibilities. In this case the AI must be limited to search only a certain number of moves in advance before making a choice. In Section \ref{Performance} the performance of the example case studies will be examined and optimisations to their implementation, as well as the performance optimisations of the library, will be explained in detail.\\
\newpage 
This thesis starts in Chapter \ref{HaskellIntroduction} with a basic introduction into the programming language Haskell for readers that are not familiar with it. Chapter \ref{SelectionMonad} will explain the concept of the novel selection monad and how this monad can be used to implement AIs for sequential games. Chapter \ref{Library} then explains the library that was made, and Chapter \ref{ExampleGames} introduces three example games that are used as case studies. The performance of these case studies are then discussed further in Chapter \ref{Performance}. The paper concludes with the discussion and conclusion in Chapter \ref{Discussion}.	

%%%%%%%%%%%%%%%%%%%%%%%%%%%%%%%%%%%%%%%%%%%%%%%%%%%%%%%%%%%%%%%%%%%%
% Einleitung
%%%%%%%%%%%%%%%%%%%%%%%%%%%%%%%%%%%%%%%%%%%%%%%%%%%%%%%%%%%%%%%%%%%%

%% 7-8 Pages
\chapter{Introduction into Haskell}\label{HaskellIntroduction}
\doublespacing
The example case studies of this project are implemented in the programming language Haskell.
Haskell is a pure functional programming language, which enables an elegant implementation of the novel \textit{selection monad} and its functionality, as well as declarative, well structured, and modular example implementations of sequential games. The following Chapter is a brief introduction to the basic concepts of Haskell to support the understanding of the following Chapters. Section \ref{Monads} introduces the concept of the monad data type and its implementation in Haskell.
\section{Haskell is a Functional Programming Language}
As a pure functional programming language, Haskell enables modular function definitions. A Haskell program consists of small function definitions that are combined to bigger functions, which eventually form a whole program. A function in Haskell consists of an arbitrary number of input parameters and computes the output. The definition of "pure" in "pure functional programming language" forbids the presence of state and side effects and, therefore, functions in Haskell cannot have side effects. The result of a function, therefore, only depends on its input parameters. Side effects and state dependent computations are not supported and are modelled with the help of monads, which will be explained further in Section \ref{Monads}.\newline
Haskell is also a strongly typed language, having a strong differentiation between data types. Every function has a specific data type describing the input parameters and the output. 

\subsection{Type System}
Haskell supports the commonly used data types: Integer (Int), Floating-point (Float), Boolean (Bool), String (String), Character (Char), among others. The type of a function will be annotated above the actual function definition with the following syntax: \\
\begin{figure}[h!]
\begin{lstlisting}[language=Haskell, style=myHaskellStyle]
functionName :: Int -> Int -> Bool
functionName param1 param2 = <functionDefinition>
\end{lstlisting}
\caption{Example function definition}
\end{figure}
\subsubsection{Type Variables}
To enable more general functions that work for more then one specific type, Haskell has the concept of type variables. Instead of writing a specific type, type variables are used in the type definition of the function. The following example function type signature is of a function that works for any list of elements with type \textbf{a} and returns the length of the list:\\
\begin{figure}[h!]
\begin{lstlisting}[language=Haskell, style=myHaskellStyleInline]
length :: [a] -> Int
\end{lstlisting}
\end{figure}
\subsubsection{Higher Order Functions}
Furthermore, Haskell also supports higher order functions. This class of functions take another function as parameter. A common example is the map function on lists, which applies a function to each element in a given list, returning a new list with the modified elements. The map function has the following type signature:\\
\begin{figure}[h!]
\begin{lstlisting}[language=Haskell, style=myHaskellStyleInline]
map :: (a -> b) -> [a] -> [b]
\end{lstlisting}
\end{figure}
\subsubsection{Custom Data Types}
In addition to the build-in types, custom data types can be defined. The following definition defines a tree data type. An object of this data type is either a Leaf or a Node with two child trees. \\
\begin{figure}[h!]
\begin{lstlisting}[language=Haskell, style=myHaskellStyleInline]
data Tree = Node Tree Tree | Leaf
\end{lstlisting}
\end{figure}\\
These custom data types can also contain already defined data types. We can therefore modify this definition so that each Node and Leaf also hold an Integer value:\\
\begin{figure}[h!]
\begin{lstlisting}[language=Haskell, style=myHaskellStyleInline]
data Tree = Node Int Tree Tree | Leaf Int
\end{lstlisting}
\end{figure}\\
Furthermore, it also supports Type variables, resulting in a tree that can contain elements of any data type:\\
\begin{figure}[h!]
\begin{lstlisting}[language=Haskell, style=myHaskellStyleInline]
data Tree a = Node a Tree Tree | Leaf a
\end{lstlisting}
\end{figure}
\subsubsection{Type classes}
Types can be grouped into type classes. A type class describes a common behaviour that all types in this type class share with each other. For example, all types that have an equality relation belong to the \textbf{Eq} type class. For each type in this type class, the equality function (==) is implemented and can be used. It is possible to restrict general functions with type variables to certain type classes. The following example is from a function that searches a list of \textbf{a}'s for a given element of type \textbf{a} and returns true if the list contains this element. To do so, it needs to call the equality function (==). It, therefore, only works for types that are in the \textbf{Eq} type class.\\
\begin{figure}[h!]
\begin{lstlisting}[language=Haskell, style=myHaskellStyleInline]
contains :: (Eq a) => [a] -> a -> Bool
\end{lstlisting}
\end{figure}\\
Other common type classes are Ord for types that can be ordered, or Show for types that can be transformed to a string.
\subsection{Functions}
In Haskell functions are the main language feature. Even values are represented as nullary functions, which is a function without input. A typical function definition consists of its type definition and the actual function body.\\
\begin{figure}[h!]
\begin{lstlisting}[language=Haskell, style=myHaskellStyleInline]
functionName :: Type1 -> Type2 -> Type3
functionName param1 param2 = <functionDefinition>
\end{lstlisting}
\end{figure}\\
The \textit{functionDefinition} part will be replaced by an expression that evaluates to a value of the corresponding output type for the given input parameters. Haskell does not support imperative computations or loops. Instead of loops, recursion is used for repeating a computation several times. Haskell also does not differ between operators and functions, therefore these operators are simple functions that are used infix. \newline
The syntax for function applications is: first the function name and then its parameters. For example, the \textit{add} Function with the following type:\\
% TODO  Alternative zu imperative computations
\begin{figure}[h!]
\begin{lstlisting}[language=Haskell, style=myHaskellStyleInline]
add :: Int -> Int -> Int
\end{lstlisting}
\end{figure}\\
can be applied as the following, to perform 2 + 2:\\
\begin{figure}[h!]
\begin{lstlisting}[language=Haskell, style=myHaskellStyleInline]
four = add 2 2
\end{lstlisting}
\end{figure}\\
This expression can also be written with the \textbf{+} function, the infix variant of the add function:\\
\begin{figure}[h!]
\begin{lstlisting}[language=Haskell, style=myHaskellStyleInline]
four = 2 + 2
\end{lstlisting}
\end{figure}
\subsubsection{Currying}
The concept of currying describes a partial application of a function. It is possible in Haskell to apply a function only to a part of its parameters, resulting in a new function that is still waiting for the rest of the parameters before evaluating. For example, a new function can be built that always adds one to a given Integer by partially applying the previously introduced add function to 1:\\
\begin{figure}[h!]
\begin{lstlisting}[language=Haskell, style=myHaskellStyleInline]
add1 :: Int -> Int
add1 = add 1
\end{lstlisting}
\end{figure}
\subsubsection{Lambda Expressions}
Haskell supports the concept of lambda expressions to define an anonymous function inside an expression, rather then defining and naming the function outside the expression. The syntax of lambda expressions is the following:\\
\begin{figure}[h!]
\begin{lstlisting}[language=Haskell, style=myHaskellStyleInline]
(\x -> x + 1)
\end{lstlisting}
\end{figure}\\
and can be used, for example, inside the application of the \textit{map} function:\\
\begin{figure}[h!]
\begin{lstlisting}[language=Haskell, style=myHaskellStyleInline]
listPlus1 = map (\x -> x + 1) [1,2,3,4]
\end{lstlisting}
\end{figure}\\
which adds one to each element in the list, resulting in the list \lstinline[language=Haskell, style=myHaskellStyleInline]{[2,3,4,5]}
In contrast, the same definition without a lambda expression, using the previous defined \textit{add1} function, could look like this:\\
\begin{figure}[h!]
\begin{lstlisting}[language=Haskell, style=myHaskellStyleInline]
listPlus1 = map add1 [1,2,3,4]
\end{lstlisting}
\end{figure}
\subsubsection{Pattern Matching}
Certain container data types, like lists or tuples, as well as custom data types like the previous defined tree data type, can be de-constructed using pattern matching. This can be done directly at the function definition, for example with the Tree data type:\\
\begin{figure}[h!]
\begin{lstlisting}[language=Haskell, style=myHaskellStyle]
isLeaf :: Tree -> Bool
isLeaf (Node t1 t2) = False
isLeaf (Leaf)	    	 = True
\end{lstlisting}
\caption{Pattern matching in the parameter definition}
\end{figure}\\
Here the deconstruction happens instead of naming the input parameter. Depending on the structure of the data type, the corresponding expression will be evaluated. As an alternative, the case keyword can be used inside an expression:\\
\begin{figure}[h!]
\begin{lstlisting}[language=Haskell, style=myHaskellStyle]
isLeaf :: Tree -> Bool
isLeaf t = case (t) of
		(Node t1 t2) 	-> False
		(Leaf)		    -> True
\end{lstlisting}
\caption{Pattern matching with the case statement}
\end{figure}
\subsection{Modules}
To organise a Haskell program, data types and collections of functions can be summarised into modules, which can then be imported into other Haskell programs. A module can be defined with the following syntax at the beginning of a file:\\
\begin{figure}[h!]
\begin{lstlisting}[language=Haskell, style=myHaskellStyle]
module ModuleName (function1, function2) where 
	function1 :: Type1 -> Type2
	function1 = ...
	
	function2 :: Type2 -> Type3
	function2 = ...
	
	helperFunction :: Type1 -> Type3
	helperFunction = ...
\end{lstlisting}
\caption{Example module definition}
\end{figure}\\
The functions in the parentheses are the functions that will be exported and made available when importing this module. On contrast, the \textit{helperFunction} will not be made available.
The module \textit{Prelude} is the main Haskell module providing the basic data types like \textbf{Int} or \textbf{Bool} and basic functions like \textbf{+} or \textbf{map}. 
\section{Monads}\label{Monads}
A monad is special type class of the previous introduced type classes. The concept of a monad originates in the mathematical field of the category theory. There are many ways to explain a monad. The following section will only give a brief introduction. For more details, there are many good tutorials out there explaining monads \cite{RealWorldHaskell}.\\
Monads can be used in many different use-cases. The most popular ones are for modelling state-full behaviour or doing Input/Output actions. To understand the concept of monads, consider the following example: We have a collection of functions that are all taking one argument as input and produce an optional outcome:\\
\begin{figure}[h!]
\begin{lstlisting}[language=Haskell, style=myHaskellStyleInline]
function1 :: a -> Maybe b
\end{lstlisting}
\end{figure}\\
The optional outcome will be modelled with the \textit{Maybe} data type which is either the result or nothing:\\
\begin{figure}[h!]
\begin{lstlisting}[language=Haskell, style=myHaskellStyleInline]
data Maybe a = Just a | Nothing
\end{lstlisting}
\end{figure}\\
A possible example function with this behaviour can be a function that searches a list for a particular element and then returns this element if the list contains it, or returns nothing if it is not in the list. \\
We now want to chain the functions so that the output of the previous function is the input of the next function. When doing this without the help of the monadic behaviour, we would need to unpack each output first and handle the \emph{Nothing} case properly, resulting in overly-nested code. 
To avoid this, the following operator would be helpful to take the two functions that need to be chained, as well as the original input: \\
\begin{figure}[h!]
\begin{lstlisting}[language=Haskell, style=myHaskellStyle]
(>>?) :: a -> (a -> Maybe b) -> (b -> Maybe c) -> Maybe c
(>>?) x f1 f2 = case f1 x of
                  Nothing  -> Nothing
                  (Just a) -> f2 a
\end{lstlisting}
\caption{Example function for chaining two Maybe operations}
\end{figure}\\
Now, it is possible to chain two functions that produce a Maybe. In order to make this function more general, the type can be reduced to:\\
\begin{figure}[h!]
\begin{lstlisting}[language=Haskell, style=myHaskellStyle]
(>>=) :: Maybe a -> (a -> Maybe b) -> Maybe b
(>>=) x f = case x of
              Nothing  -> Nothing
              (Just a) -> f a
\end{lstlisting}
\caption{Definition of the monadic bind operator}
\end{figure}\\
Compared to the type class definition of monads, the type signature is similar:\\
\begin{figure}[h!]
\begin{lstlisting}[language=Haskell, style=myHaskellStyle]
class Monad m where
    (>>=)  :: m a -> (a -> m b) -> m b
    return :: a -> m a
\end{lstlisting}
\caption{Definition of the monad type class}
\end{figure}\\
So, in general, monads are container data types that surround the actual data. If there are a bunch of functions all producing the same monadic output, the monadic \textit{$>>$=} function provides a way of chaining them without the need to manually unpack them. 

%%%%%%%%%%%%%%%%%%%%%%%%%%%%%%%%%%%%%%%%%%%%%%%%%%%%%%%%%%%%%%%%%%%%
% Einleitung
%%%%%%%%%%%%%%%%%%%%%%%%%%%%%%%%%%%%%%%%%%%%%%%%%%%%%%%%%%%%%%%%%%%%
% 8 Pages
\chapter{Selection Monad}\label{SelectionMonad}
\doublespacing
The selection monad was first described in the context of finding optimal plays in sequential games by Martín Escardó and Paulo Oliva in 2010 \cite{SequtentialGames}. This section is an introduction to the selection monad, and how it can be used to find optimal plays and strategies in sequential games.  
\section{Selection Functions}\label{SelectionFunctions}
To understand the selection monad, it is helpful to understand selection functions first. Selection functions are functions that select an element out of a given list of elements based an internal evaluation of the elements. An example selection function could be a function that takes a list of strings as an input and searches the whole list for a string with the length 3. If the list contains such a string, one of these strings will be the output. If not, a random string will be selected. An implementation of this function can be seen in Figure \ref{SelectionFunction1}.
\begin{figure}[h!]
\begin{lstlisting}[language=Haskell, style=myHaskellStyleInline]
selectionFunction :: [String] -> String
selectionFunction [x]    = x
selectionFunction (x:xs) = if length x == 3 then x else selectionFunction xs
\end{lstlisting}
\caption{Example selection function on strings}\label{SelectionFunction1}
\end{figure}\\
This implementation is very specific, only works for strings and the selection function is embedded inside the function definition. To generalise this definition, type variables and higher order functions can be used, to define a more general version that works for every type and for many selection behaviours. A more general implementation could be the following:
\begin{figure}[h!]
\begin{lstlisting}[language=Haskell, style=myHaskellStyle]
selectionFunction2 :: [a] -> (a -> Bool) -> a
selectionFunction2 [x] f    = x
selectionFunction2 (x:xs) f = if f x then x else selectionFunction2 xs f
\end{lstlisting}
\caption{More general definition for a selection function}
\end{figure}\\
This implementation is still specific to selection functions that only work with booleans. To extend this to a function that works with a more general type of truth values \textit{r}, parts of the selection behaviour need to be embedded in the function definition. For example we have a function that maps every value in the input list to an integer value and we want to select the value that is assigned the highest mapping. A possible implementation could look like this:
\begin{figure}[h!]
\begin{lstlisting}[language=Haskell, style=myHaskellStyle]
selectionFunctionMax :: (Ord r) => [a] -> (a -> r) -> a
selectionFunctionMax [] f = undefined
selectionFunctionMax xs f = last (sortOn f xs)
\end{lstlisting}
\caption{Definition for a maximum selection function}
\end{figure}\\
\textit{Last} and \textit{sortOn} are build-in functions of the Haskell prelude module. \textit{SortOn} sorts a list with respect to a given function, while \textit{last} returns the last element of a list. Note, this achieves a more general type definition with the cost of loosing generality in the function implementation itself. \\
With this in mind, a new general type for selection functions can be defined. This is done using custom data types and type variables. Here the type variable \textit{r} stands for a general type of truth values and \textit{x} denotes the type variable of the list elements. The type and the adjusted type signature of the maximum selection function can be defined as below:
\begin{figure}[h!]
\begin{lstlisting}[language=Haskell, style=myHaskellStyle]
type J r x = (x -> r) -> x

selectionFunctionMax2 :: (Ord b) => [a] -> J b a
\end{lstlisting}
\caption{Type definition of the selection function type J}
\end{figure}
\subsection{Quantifier functions}
% TODO: deutlicher erklären mehr beispiel
Closely related to this new selection type is the quantifier type also described by Martín Escardó and Paulo Oliva in their paper from 2010 \cite{SequtentialGames}. This quantifier type returns the result of the selection function, instead of the element that is selected. For example, in the context of the selection function that is looking for a string with the length of 3, the corresponding quantifier would return true if such a string exists and false if not. Because of this close relationship, an overline function can be defined which converts the selection type to the quantifier type. The implementation could look like this:
\begin{figure}[h!]
\begin{lstlisting}[language=Haskell, style=myHaskellStyle]
type K r x = (x -> r) -> r

overline :: J r x -> K r x
overline e p = p (e p)
\end{lstlisting}
\caption{Type definition of the quantifier data type. Basically the evaluation function is applied again to the result of the selection function.}
\end{figure}
\subsection{Modelling sequential games}
In order to bring selection functions in the context of implementing an AI for sequential games, we can model the way the AI decides which move it wants to play as a selection function. This selection function is selecting the best possible move out of a list of all possible moves. The type signature of this function could look like this:\\
\begin{figure}[h!]
\begin{lstlisting}[language=Haskell, style=myHaskellStyle]
bestMove :: [Move] -> J r Move
bestMove' :: [Move] -> (Move -> r) -> Move
\end{lstlisting}
\caption{Functions calculating the best move. Both have the same type but the first uses the previously defined custom data type J}
\end{figure}\\
Here \textit{r} is describing the current state of the game, for example if the player is winning or loosing after this move, and \textit{Move} is a custom data type that is representing a move in an arbitrary game. Furthermore a whole play of a sequential game can be modelled as a list of all \textit{bestMove} selection functions:
\begin{figure}[h!]
\begin{lstlisting}[language=Haskell, style=myHaskellStyle] 
sequentialPlay :: [[Move] -> J r Move]]
\end{lstlisting}
\end{figure}\\
Martín Escardó and Paulo Oliva introduced the \textit{bigotimes} function in their paper from 2010 \cite{SequtentialGames}. This function combines a list of selection functions into one big selection function. This combined selection function can be seen as the function selecting a list of optimal moves that represent a perfect play. 
\begin{figure}[h!]
\begin{lstlisting}[language=Haskell, style=myHaskellStyle]
otimes :: J r x -> (x -> J r [x]) -> J r [x]
otimes e0 e1 p = a0 : a1
  where a0 = e0(\x0 -> overline (e1 x0) (\x1 -> p(x0:x1)))
        a1 = e1 a0 (\x1 -> p(a0 : x1))
        overline e p = p(e p)

bigotimes :: [[x] -> J r x] -> J r [x]
bigotimes [] = \p -> []
bigotimes (e : es) = e [] `otimes` (\x -> bigotimes[\xs->d(x:xs) | d <- es])
\end{lstlisting}
\caption{\textit{bigotimes} function definitions taken from \cite{SequtentialGames,SelectionMonadCode}} \label{BigotimesImplementation}
\end{figure}\\
The implementation of this bigotimes function as shown in Figure \ref{BigotimesImplementation} is explained in detail in Martín Escardó and Paulo Olivas paper from 2010 \cite{SequtentialGames}. A detailed explanation of this function will be omitted in this paper due to the high complexity of these functions and the limited space in this paper. 
\section{Implementing Games}\label{ImplementingGames}
With the help of this \textit{bigotimes} function, it is possible to define functions that calculate an optimal play, the outcome of an optimal play and the optimal strategy. When implementing a sequential game, the programmer needs to define two essential functions. First a function that evaluates a given list of moves to a value that represents the state of the game. Here often integer values are used where 1 stands for player 1 wins, 0 stands for a draw and -1 stands for player 2 wins. This function will be named \textit{p} in the following case study implementations. The type signature of this function looks like this, where \textit{R} represents the state of the game: 
\begin{figure}[h!]
\begin{lstlisting}[language=Haskell, style=myHaskellStyleInline]
p :: [Moves] -> R
\end{lstlisting}
\caption{p type definition}
\label{p}
\end{figure}\\
The second function represents the strategies that both players use to play the game. It is a list of selection functions that select one move out of all possible moves. In the following case studies this function will be called \textit{epsilons} and the type signature of this function is the following:
\begin{figure}[h!]
\begin{lstlisting}[language=Haskell, style=myHaskellStyleInline]
epsilons :: [[Move] -> J R Move]
\end{lstlisting}
\caption{epsilons type definition}
\label{epsilons}
\end{figure}\\
Having the \textit{p} and \textit{epsilons} functions defined, the following three functions can be called to calculate the optimal play as defined in Figure \ref{OptimalPlayCode}.
\begin{figure}[h!]
\begin{lstlisting}[language=Haskell, style=myHaskellStyleInline]
optimalPlay :: ([b] -> a) -> [[b] -> J a b] -> [b]
optimalPlay p epsilons = bigotimes epsilons p

optimalOutcome :: ([b] -> a) -> [[b] -> J a b] -> a
optimalOutcome p epsilons = p $ optimalPlay p epsilons

optimalStrategy :: ([b] -> a) -> [[b] -> J a b] -> [b] -> b
optimalStrategy p epsilons as = head(bigotimes epsilons p')
   where p' xs = p (as ++ xs)
\end{lstlisting}
\caption{Definitions calculating a perfect play}
\label{OptimalPlayCode}
\end{figure}\\
In order to implement interactive games where the AI only chooses one move instead of the whole optimal play, the optimal strategy function calculates the whole optimal play but only returns the first move out of this optimal play list. It also has an additional parameter to take previous made moves into account. 
\subsection{Monadic selection function and bigotimes function}
This previously defined custom data type for selection functions can now be made an instance of the monad type class. \\
This enables an easier definition of the \textit{bigotimes} function. A detailed explanation of the instantiation of the selection monad, and of the \textit{bigotimes} function can be found in the original paper from Martín Escardó and Paulo Oliva \cite{SequtentialGames}. The corresponding implementation can be found in Figure \ref{SelectionMonadCode} and Figure \ref{bigotimesCode}.\\
\begin{figure}[h!]
\begin{lstlisting}[language=Haskell, style=myHaskellStyle]
newtype J r x = J {selection :: (x -> r) -> x}

instance Monad (J r) where
  return = unitJ
  e >>= f = muJ(functorJ f e)

morphismJK :: J r x -> K r x
morphismJK e = K(\p -> p(selection e p))

unitJ :: x -> J r x
unitJ x = J(\p -> x)

functorJ :: (x -> y) -> J r x -> J r y
functorJ f e = J(\q -> f(selection e (\x -> q(f x))))

muJ :: J r(J r x) -> J r x
muJ e = J(\p -> selection(selection e(\d -> quantifier(morphismJK d) p)) p)
\end{lstlisting}
\caption{Selection monad definition. Taken from \cite{SequtentialGames,SelectionMonadCode}}
\label{SelectionMonadCode}
\end{figure}
\begin{figure}[h!]
\begin{lstlisting}[language=Haskell, style=myHaskellStyle]
otimes :: Monad m => m x -> (x -> m y) -> m(x, y)
xm `otimes` ym =
    do x <- xm
       y <- ym x
       return (x, y)

bigotimes :: Monad m => [[x] -> m x] -> m[x]
bigotimes [] = return []
bigotimes (xm : xms) =
    do x <- xm []
       xs <- bigotimes[\ys -> ym(x:ys) | ym <- xms]
       return(x : xs)
\end{lstlisting}
\caption{\textit{bigotimes} function definitions using monadic behavour. Taken from \cite{SequtentialGames,SelectionMonadCode}}
\label{bigotimesCode}
\end{figure}\\

%%%%%%%%%%%%%%%%%%%%%%%%%%%%%%%%%%%%%%%%%%%%%%%%%%%%%%%%%%%%%%%%%%%%
% Einleitung
%%%%%%%%%%%%%%%%%%%%%%%%%%%%%%%%%%%%%%%%%%%%%%%%%%%%%%%%%%%%%%%%%%%%
% 6-8 Pages

\doublespacing
\chapter{Library}\label{Library}
The previous Chapter \ref{SelectionMonad} introduced functions that calculate the optimal play, as well as the monadic definition of the selection data type and the quantifier data type, which are summarised in the library. The goal of this library is to support the implementation of AIs in sequential games. The full implementation of the library can be found in the supporting material. The library was built by analysing previously existing case studies \cite{SelectionMonadCode}, and with the experience that was gained by implementing the example games introduced in Chapter \ref{ExampleGames}. The example games were then edited to work with the library and, therefore, the library is introduced before the example game implementations.
The following Chapter explains the library's structure and design, and also gives an overview on how to use the library to implement AIs for sequential games. Some parts of the library are also tested using unit tests, which will be discussed in Chapter \ref{Performance}.
\section{Structure}
The library consists of five different sub-modules, each containing a different aspect of the selection monad. The first two modules contain the definition and type class instantiation for the selection monad and the quantifier monad as defined in Figure \ref{SelectionMonadCode}. The next module contains the \textit{bigotimes} function definition as introduced in Figure \ref{bigotimesCode}. Another module contains the optimal play functions from Figure \ref{OptimalPlayCode}. The final module provides some minimum and maximum functions that can be used to build the function that were introduced in Figure \ref{epsilons}. These minimum and maximum functions will be discussed further in Section \ref{MinMax}.\\
In order to use the library, a potential user needs to implement two functions that where previously mentioned in Chapter \ref{SelectionMonad}: The \textit{epsilons} and \textit{p} function. Before implementing these two functions, at least two data types need to be defined: The \textbf{Move} data type, which represents a move in the game, and the \textbf{R} data type, which represents the outcome of the game. In an simple game, the outcome of the game can be described as either a win, lose, or a draw, but the library also supports more sophisticated outcome types. Which types the library supports will be discussed in Section \ref{MinMax}. With these data type definitions, the \textit{p} and \textit{epsilons} function can now be defined. \textit{P} is the function which calculates the outcome of the game for a given list of previous moves. \textit{Epsilons} is a list of selection functions, modelling the strategy which optimal players would use to make their move. To support the implementation of the \textit{epsilons} function, the library provides a set of minimum and maximum functions that are already selection functions.
\section{Different min max functions}\label{MinMax}
In game theory, the minimax algorithm is used to calculate optimal plays. With the help of the selection monad and the \textit{bigotimes} function, it is easy to define the optimal play functions as introduced in Figure \ref{OptimalPlayCode}. It is essentially an implementation of the minimax algorithm. The algorithm explores all possible moves and calculates the best move for each player at each stage. In this algorithm, the sequence of all possible moves of the game are represented as a tree. The algorithm explores the whole tree from the bottom up and assigns each leaf a value. This value represents the outcome of the game based on all previous played moves. Then, each parent node will be assigned the value out of all child node values that is the best for the current player. As one player tries to minimise the outcome and the other tries to maximise the outcome, minimum and maximum functions are used to find the optimal outcome. Figure \ref{MinMaxTree} shows an example tree of the minimax algorithm.\\
\begin{figure}[h!]
		\centering
    	\begin{tikzpicture}[->,>=stealth',level/.style={sibling distance = 4.7cm/#1,level distance = 1.25cm}] 
		\node [arn_r] {2}
        	child{ node [arn_n] {-3} 
            	child{ node [arn_r] {-3} 
            		child{ node [arn_n] {-5}}
					child{ node [arn_n] {-3}}}
    			child{ node [arn_r] {5}
    				child{ node [arn_n] {5}}
					child{ node [arn_n] {2}}}}
    		child{ node [arn_n] {2}
            	child{ node [arn_r] {2}
            		child{ node [arn_n] {0}}
					child{ node [arn_n] {2}}}
            	child{ node [arn_r] {4}
            		child{ node [arn_n] {-3}}
					child{ node [arn_n] {4}}}}; 
		\end{tikzpicture}
		\caption{This three represents a small example game where the red player wants to maximise the outcome and the black player wants to minimise it}
		\label{MinMaxTree}
\end{figure}
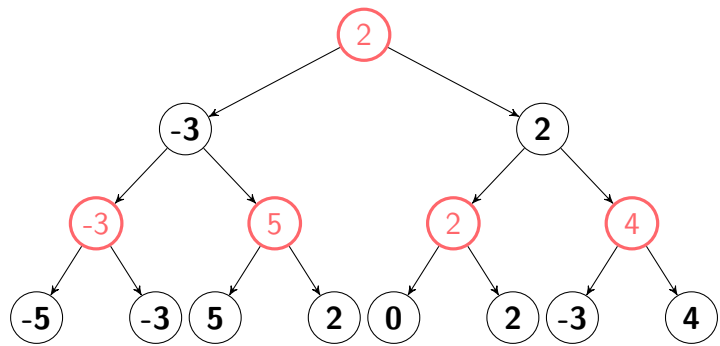\\
The library provides a set of different minimum and maximum functions that are already using the selection monad and are suitable for different ways of outcome types. These functions select the value of a given list, producing the highest outcome of a given outcome function. In the context of implementing games, the type \textit{a} is representing a move and \textit{b} is representing the outcome of the game:
\begin{figure}[h!]
\begin{lstlisting}[language=Haskell, style=myHaskellStyle]
epsilonMax :: (Ord b) => [a] -> J b a
epsilonMax' :: (Ord b) => [a] -> (a -> b) -> a
\end{lstlisting}
\caption{Two type definitions of a potential maximum function. The first one uses the selection monad. Both type definitions express the same functionality.}
\end{figure}
\newpage
\subsection{Different ways of modelling the outcome}
There are different ways of modelling the outcome of a game. Some require special minimum and maximum functions. For performance reasons, the library supports four different ways of modelling the outcome. 
\subsubsection{Generic}
The easiest way of modelling the outcome is to use Integer or Float values. In this case, a high value means that the first player wins and a low value that the second player wins instead. The implementation of the minimum and maximum functions in this case is straight forward:
\begin{figure}[h!]
\begin{lstlisting}[language=Haskell, style=myHaskellStyle]
epsilonMin :: (Ord b) => [a] -> J b a
epsilonMin xs = J(epsilonMin' xs)

epsilonMin' :: (Ord b) => [a] -> (a -> b) -> a
epsilonMin' [] _ = undefined
epsilonMin' xs f = head $ sortOn f xs
\end{lstlisting}
\caption{Generic minimum implementation}
\end{figure}
\subsubsection{1,0,-1}
Another way is to use integers, but only use the three numbers 1, 0 and -1. Here, 1 means player one wins, 0 that the game is a draw, and -1 that player two wins. Because of the limited upper and lower bound, it is possible to optimise the algorithm and stop searching when the maximum or minimum is found. The implementation of this algorithm looks like this:
\begin{figure}[h!]
\begin{lstlisting}[language=Haskell, style=myHaskellStyleInline]
epsilonMinThree :: [a] -> J Three a
epsilonMinThree xs = J(epsilonMin' xs)

epsilonMinThree' :: [a] -> (a -> Three) -> a
epsilonMinThree' [] _     = undefined
epsilonMinThree' [x] _    = x
epsilonMinThree' (x:xs) f | f x == 1 = epsilonMinThree' xs f
                          | f x == -1  = x
                          | otherwise = case findMin xs f of
                                          (Just y) -> y
                                          Nothing  -> x
    where findMin [] _     = Nothing
          findMin (x:xs) f = if f x == -1 then Just x else findMin xs f
\end{lstlisting}
\caption{Minimum implementation for the (1,0,-1) type}
\end{figure}\\
\subsubsection{Bool}
Instead of three values, it is also possible to have just two values representing the outcome of the game. Here, a boolean is used to represent the game's outcome. The following implementation is also optimised to stop searching as soon as the minimum/maximum is found:\\
\begin{figure}[h!]
\begin{lstlisting}[language=Haskell, style=myHaskellStyleInline]
epsilonMinBool :: [a] -> J Bool a
epsilonMinBool xs = J(epsilonMinBool' xs)

epsilonMinBool' :: [a] -> (a -> Bool) -> a
epsilonMinBool' [] _     = undefined
epsilonMinBool' [x] _    = x
epsilonMinBool' (x:xs) f = if not $ f x then x else epsilonMinBool' xs f
\end{lstlisting}
\caption{Bool minimum implementation}
\end{figure}
\subsubsection{Tuple}
In many cases, it is not enough to only represent the outcome of the game, but it also necessary to track after how many moves this outcome is reached. For example, even if a player is losing with every move possible to make, they still want to play as long as possible in case the opponent makes a mistake. This outcome type is often necessary when implementing an interactive game where a real person plays against the computer. To model this outcome type, a tuple consisting of two integers is used. The first represents the outcome and the second the number of moves it takes to achieve this outcome. The first value must follow the format that positive values are used when Player One wins and negative values are used when Player Two wins. The function selects the shortest way if it has a winning strategy; otherwise, it tries to stay as long as possible in the game. An implementation could look like this:
\begin{figure}[h!]
\begin{lstlisting}[language=Haskell, style=myHaskellStyleInline]
epsilonMinTuple :: [a] -> J (Int, Int) a
epsilonMinTuple xs = J(epsilonMinTuple' xs)

epsilonMinTuple' :: [a] -> (a -> (Int, Int)) -> a
epsilonMinTuple' [] _ = undefined
epsilonMinTuple' xs f = let list = sortOn fst (map (\x -> (f x, x)) xs) in
                            if  fst (fst $ head list) > 0
                            then snd $ last $ filter (\x -> fst (fst x) == fst (fst $ head list)) list
                            else snd $ head list
\end{lstlisting}
\caption{Tuple minimum implementation}
\end{figure}
\subsection{Parallel versions}
For the generic and tuple version, the performance of the whole algorithm can be improved by executing the outcome function in parallel. The implementations of these versions first apply the outcome function to the whole list of elements before making a decision, where the other versions stop searching when the maximum is found. These parallel versions have significant effects on the performance and cut down the calculation time, in some cases by a factor of 100. The implementation is done with the parallel library, which is one of the standard Haskell libraries. The implementations of these parallel functions look like this:
\begin{figure}[h!]
\begin{lstlisting}[language=Haskell, style=myHaskellStyleInline]
epsilonMinParalell ::(NFData a, NFData b, Ord a, Ord b) => [b] -> J a b
epsilonMinParalell xs = J(epsilonMinParalell' xs)

epsilonMinParalell' :: (NFData a, NFData b, Ord a, Ord b) => [a] -> (a -> b) -> a
epsilonMinParalell' xs f = snd $ minimum $ parMap rdeepseq (\x -> (f x, x)) xs

epsilonMinTupleParalell :: (NFData a) => [a] -> J (Int, Int) a
epsilonMinTupleParalell xs = J(epsilonMinTupleParalell' xs)

epsilonMinTupleParalell' :: (NFData a) => [a] -> (a -> (Int, Int)) -> a
epsilonMinTupleParalell' [] _ = undefined
epsilonMinTupleParalell' xs f = let list = sortOn fst (parMap rdeepseq (\x -> (f x, x)) xs) in
                            if  fst (fst $ head list) > 0
                            then snd $ last $ filter (\x -> fst (fst x) == fst (fst $ head list)) list
                            else snd $ head list
\end{lstlisting}
\caption{Parallel implementations of minimum functions}
\end{figure}
\subsection{Usage of the minimum and maximum functions}
These minimum and maximum functions can be used to define the \textit{epsilons} function that is necessary for the optimal play computation. As introduced in Chapter \ref{SelectionMonad}, a classical two-player game can be modelled as a list of decisions where each player decides their next move based on the previous moves. An infinite list is built to  alternate between minimum and maximum selection functions. In order to limit the search depth, only a certain number of elements are taken from the beginning of the list. A possible \textit{epsilons} implementation that searches the tree of all possible moves to the depth of 9 and uses the previously introduced tuples outcome could look like this:
\begin{figure}[h!]
\begin{lstlisting}[language=Haskell, style=myHaskellStyleInline]
epsilons :: [[Move] -> J R Move]
epsilons = take 9 all
  where all = epsilon1 : epsilon2 : all
        epsilon1 history = epsilonMaxTupleParalell (getPossibleMoves history)
        epsilon2 history = epsilonMinTupleParalell (getPossibleMoves history)
\end{lstlisting}
\caption{Example epsilons function using the tuple minimum and maximum function}
\end{figure}

%%%%%%%%%%%%%%%%%%%%%%%%%%%%%%%%%%%%%%%%%%%%%%%%%%%%%%%%%%%%%%%%%%%%
% Einleitung
%%%%%%%%%%%%%%%%%%%%%%%%%%%%%%%%%%%%%%%%%%%%%%%%%%%%%%%%%%%%%%%%%%%%
% 12 pages

\chapter{Example Game Implemenations}\label{ExampleGames}
This Section introduces the implementations of the three example games used as case studies to explore the selection monad's ability to support AI for sequential games. All case studies are implemented in the programming language Haskell. A brief introduction to Haskell can be found in Chapter \ref{HaskellIntroduction}. The following sections introduce the rules of each of the games briefly. Then, the implementation details of each game will be presented and finally, the suitability of the selection monad will be discussed. Each of the case studies needed to deal with the complexity of the corresponding game - this complexity and the corresponding performance optimisations will be discussed separately in Chapter \ref{Performance}.
\section{Connect Four} % Todo mentione the interactive version
Connect Four is a sequential two-player game. It is played on a 7x6 grid board. A turn consists of a player choosing a column of the board and inserting a disk of his colour. The disk then falls down to the bottom of the board occupying the next available space in this column. The player who first has 4 in a row (Horizontal, Vertical, Diagonal) wins the game. \\
A perfect strategy to play this game was fist introduced by Victor Allis \cite{ConnectThreeSolution1} in 1988.
Allis performed a knowledge based approach, describing nine strategies leading to a win for the beginning player. A bruteforce approach to this problem by trying all of the the over four trillion different games \cite{ConnectFour}, was done in 1995 by John Tromp \cite{ConnectThreeSolution2}. In a perfect play, the beginning player can achieve a win within the 41st move.\\
The following section introduces two different implementations of Connect Four: one that calculates a perfect play, and another interactive version where a human player plays against an AI.
A first implementation of the game with the selection monad showed that, without any pruning (limiting of the search space of all possible moves), it is not feasible to calculate a perfect strategy for Connect Four because of the huge amount of different games that are needed to be explored. More details on the performance of these implementations are discussed in Chapter \ref{Performance}. In order to reduce the number of possible moves, the game was simplified to a Connect Three version where the board size was reduced to 5x3 (4x3 for the interactive version), and to win the game it is only necessary to have three in a row. For this simplified version, it was possible to compute a perfect play in under a minute. The following section will introduces the implementation details. 
\subsection{Connect Three}
In order to implement Connect Three, first four data types need to be defined. The board is represented as a matrix using a third party matrix library developed by Daniel Díaz \cite{MatrixLibrary}.
This library enables constant access times and is used instead of nested lists in order to improve the performance. The board is then filled with values of type \textit{Player}, which are either Player 1, Player 2, or Nothing (X,O,N). A move is represented as an Integer, indicating which column a disk is inserted into. The outcome \textit{R} of a game is represented by a tuple of two Integers. The first Integer represents the state of the game; \textbf{1} for when Player 1 wins, \textbf{0} for a draw and \textbf{-1} for when Player 2 wins. The second Integer counts the number of moves that are made before the outcome is reached.
\begin{figure}[h!]
\begin{lstlisting}[language=Haskell, style=myHaskellStyleInline]
type R = (Int,Int)
type Move = Int
type Board = Matrix Player
data Player = X | O | N
\end{lstlisting}
\caption{Connect Three data types}
\end{figure}\\
Next, a \textit{wins} function needs to be defined that determines if a win was achieved for a given board and player. Therefore, the whole board is checked for the winning criteria (three in a row). The full implementation of this function can be found in the supporting material.
\begin{figure}[h!]
\begin{lstlisting}[language=Haskell, style=myHaskellStyleInline]
wins :: Board -> Player -> Bool
\end{lstlisting}
\caption{Type signature of the Connect Three win function}
\end{figure}\\
Now utility functions are defined to do the handling of the insertion of a disk into the board and computation of a list of possible moves depending on a list of previous moves. The \textit{insert} function checks a column from the bottom and inserts the player's disk at the first appearance of an empty cell (represented by an N). The \textit{getPossibleMoves} function computes all remaining possible moves by checking how many discs were already inserted into the corresponding column (It is only possible to insert 3 disks in a column).\\
\begin{figure}[h!]
\begin{lstlisting}[language=Haskell, style=myHaskellStyle]
insert :: Move -> Player -> Board -> Board
insert m = insert' (m,1)
  where insert' (x,y) p b = if b ! (x,y) == N
											then setElem p (x, y) b
											else insert' (x, y + 1) p b

getPossibleMoves :: [Move] -> [Move]
getPossibleMoves [] = [1..5]
getPossibleMoves xs = filter (\x -> length (elemIndices x xs) < 3) [1..5]
\end{lstlisting}
\caption{Connect Three utility functions}
\end{figure}\newpage
\subsubsection{Building the AI}
In order to build the AI for the Connect Three game, the \textit{p} and \textit{epsilons} functions need to be defined. These functions where introduced in Chapter \ref{SelectionMonad} and explained in detail in Chapter \ref{Library}.
To define the \textit{p} function that maps a list of previous played moves to the outcome, a helper \textit{outcome} function is defined. This \textit{outcome} function takes the starting player, a list of played moves, a board, and the count of moves that are already played on the given board, and inserts all moves from the list of moves into the given board. For each step, the function checks if this move is a winning move. In this case, it stops inserting and returns the outcome.
\begin{figure}[h!]
\begin{lstlisting}[language=Haskell, style=myHaskellStyle]
value :: Board -> Int
value b  | wins b W  = 1
         | wins b B  = -1
         | otherwise = 0
         
outcome :: Player -> [Move] -> Board -> Int -> R
outcome _ [] b i       = (value b, i)
outcome p (m : ms) b i = let nb = insert m p b in
                         		if wins nb p 
                         		then (value nb, i + 1) 
                         		else outcome (changePlayer p) ms nb (i + 1)

p :: [Move] -> R
p ms = outcome X ms (matrix 5 3 (const N)) 0
\end{lstlisting}
\caption{Connect Three outcome function}
\end{figure}\\
To model the AI's behaviour, the parallel minimum and maximum functions for tuples are used. These functions are provided by the library. The first player wants to maximise the outcome and the second player wants to minimise the outcome. An infinite list of alternating Min and Max selection functions is built. By constantly increasing the search depth, an optimal play was found after 9 moves. The corresponding \textit{epsilons} function therefore only takes the first 9 elements of this infinite list of selection functions. The \textit{epsilons} function was implemented like this:
\begin{figure}[h!]
\begin{lstlisting}[language=Haskell, style=myHaskellStyleInline]
epsilons :: [[Move] -> J R Move]
epsilons = take 9 all
  where all = epsilonX : epsilonO : all
        epsilonX h = epsilonMaxTupleParalell (getPossibleMoves h)
        epsilonO h = epsilonMinTupleParalell (getPossibleMoves h)
\end{lstlisting}
\caption{Connect Three epsilons function}
\end{figure}\newpage
Then, in the main function, the optimal game function is called with the previous \textit{p} and \textit{epsilons} function: 
\begin{figure}[h!]
\begin{lstlisting}[language=Haskell, style=myHaskellStyleInline]
main :: IO ()
main = do
  let optimalGame = optimalPlay p epsilons
  putStr ("An optimal play for " ++ gameName ++ " is "
     ++ show optimalGame
     ++ "\nand the optimal outcome is " ++ show (p optimalGame) ++ "\n")
\end{lstlisting}
\caption{Connect Three main function}
\end{figure}\\
This then calculates an optimal play where the first player wins after nine moves. The resulting play is: \lstinline[language=Haskell,style=myHaskellStyleInline]{[2,4,3,1,2,2,3,5,1]}.
\subsection{Interactive version}
In order to make an interactive version of this Connect Three game, the \textit{epsilons} function needs a small adjustment. The new \textit{epsilons} function needs to take the previously played moves into account. It also needs to prevent the search depth from becoming bigger than the number of possible rounds that are left in the game. The adjusted epsilons function can be found in Figure \ref{InteractiveConnectThree}.
Finally, a command line interface was defined to ask the human player for a move, and then to call the \textit{optimalStrategy} function to calculate the AI's move. The full implementation of this interactive version can be found in the supporting material.\\
Additionally to the interactive version of Connect Three, an interactive version of the original Connect Four was implemented. It is basically the same implementation as the Connect Three version, where the rules where modified to represent the original rules and the AI was limited to explore only six moves in advance.\\
\begin{figure}[h!]
\begin{lstlisting}[language=Haskell, style=myHaskellStyleInline]
epsilons :: [Move] -> [[Move] -> J R Move]
epsilons h = take (if (8 + length h) > (12 - length h) then 12 - length h else 9 + length h) all
  where all = epsilonO : epsilonX : all
        epsilonX history = epsilonMaxTupleParalell (getPossibleMoves (h ++ history))
        epsilonO history = epsilonMinTupleParalell (getPossibleMoves (h ++ history))
\end{lstlisting}
\caption{Connect Three interactive epsilons function}\label{InteractiveConnectThree}
\end{figure}
\subsection{Discussion}\label{ConnectThreeDiskussion}
Due to the simplification of the game rules, the calculation of a perfect play in Connect Three can be done in under five seconds. The original Connect Four version, on the other hand, still needs four minutes to compute nine moves in advance. This would result in an estimated computation time of for the whole game of $2*10^{26}$ years. A detailed discussion on how this performance was achieved, and what actions where taken to improve the performance, can be found in Chapter \ref{Performance}. Despite the performance issues for the whole game, the simplified Connect Three version shows perfectly how the selection monad enables a concise and expressive programming style. With the use of the library, the implementation of the game can be done within 65 lines of code. Furthermore, the interactive version shows that the library provides an easy way to both calculate a perfect play with only small adjustments to the code, and also a playable interactive version. It is also possible to limit the search space so that the AI is calculating to a smaller number of moves in advance. When limiting the search space, it is possible to implement an interactive version of the original Connect Four, that explores only six moves in advance, instead of exploring all possible moves. The AI in this state can already be a challenging opponent for a human player. The AI's strategy is to make random moves and prevent the human player from winning when necessary. It also takes the opportunity to make winning moves.\\
The outcome function only differentiates between an win, draw or lose. This is a sufficient outcome function when the complexity of the game allows it to explore all possible moves. When limiting the search space, a more sophisticated outcome function will probably improve the AI's behaviour. When taking the current positioning at the board into account when calculating the outcome, the AI aims to improve its current position instead of just preventing the opponent from winning, as long as it has the opportunity to achieve a win on its own. However, the interactive Connect Four version shows that it is possible to achieve good results with the simple win/draw/loose outcome function.
\section{Sudoku}
Sudoku is a logic based puzzle. The goal is to fill the gaps in a 9x9 matrix, so that each row and column and sub block do not contain any duplicates. The nine sub blocks are formed of 3x3 blocks, in which the whole matrix is divided. The number of gaps in the matrix define the difficulty of the puzzle. Solving a Sudoku is NP complete, meaning there is no algorithm to solve within polynomial time; however, there are heuristic algorithms that can efficiently solve Sudokus \cite{SolveSudoku1, SolveSudoku2}. The goal of this case study is to explore the monad's ability to solve one-player puzzles with a simple outcome function.
\subsection{Implementation}
First, the Board, Move and Outcome types are defined. The board is a 9x9 matrix of Integers using the third party library matrix library \cite{MatrixLibrary}. A move is a triple where the first two Integers are a coordinate pointing to a position in the matrix, and the third Integer is the value that should be written into this field. The outcome defines if a board is a valid solution to the Sudoku, and is therefore represented as a boolean.\\
\begin{figure}[h!]
\begin{lstlisting}[language=Haskell, style=myHaskellStyleInline]
type R = Bool
type Move = (Int, Int, Int)
type Board = Matrix Int
\end{lstlisting}
\caption{Sudoku data type definition}
\end{figure}\\
Next, the win function needs to be defined. The win function checks the whole board whether the Sudoku is still valid. Because this function is relatively time consuming to compute, and it is not necessary to check the whole board if only one move is applied to a board, a second \textit{win'} function is also defined to check, for a given move, if this move is valid on a given board. The implementation of this win function is very extensive and will be omitted because of the limited space. In can be found in the supporting material.\\
\begin{figure}[h!]
\begin{lstlisting}[language=Haskell, style=myHaskellStyleInline]
wins :: Board -> Bool
wins' :: Move -> Board -> Bool
\end{lstlisting}
\caption{Sudoku: Different type signatures of the wins functions}
\end{figure}\\
To define the \textit{p} function, an \textit{outcome} function is defined to insert a list of moves into a given board and check, for each insertion, if the board is still valid. A starting board is also defined, where some numbers are already inserted into an empty board, to form the initial puzzle. The implementation of the \textit{p} and \textit{outcome} functions look like this:
\begin{figure}[h!]
\begin{lstlisting}[language=Haskell, style=myHaskellStyleInline]
p :: [Move] -> R
p ms = outcome ms startingBoard

outcome :: [Move] -> Board -> R
outcome [] b     = wins b
outcome (x:xs) b = let nb = insert x b in if wins' x b then outcome xs nb else wins nb
\end{lstlisting}
\caption{Sudoku outcome function}
\end{figure}\newpage
To describe the AI that solves the Sudoku, the maximum function on boolean is used. The AI wants to make only moves that produce a valid Sudoku in the end. Therefore, an infinite list of selection functions is created and then reduced to a list of the same length as there are gaps in the Sudoku. The \textit{startingMoves} variable is a list of moves that contain the initially defined values that are necessary to define the puzzle. The implementation of the \textit{epsilons} function looks like this:
\begin{figure}[h!]
\begin{lstlisting}[language=Haskell, style=myHaskellStyleInline]
epsilons :: [[Move] -> J R Move]
epsilons = take 12 all
  where all = epsilon' : all
        epsilon' history = epsilonMaxBool (getPossibleMoves (startingMoves ++ history))
\end{lstlisting}
\caption{Sudoku epsilons function}
\end{figure}
\subsection{Discussion}
This case study shows that it is generally possible to solve puzzles like Sudoku with the help of the selection monad. However, the algorithm that is implemented by the library to solve this game (Minimax) is growing expectationally with each layer of moves. Therefore the case study implementation is able to solve Sudokus with 12 gaps in about 100 seconds, but every additional gap increases the computation time by a factor of approximately 3. A detailed discussion of the performance of this implementation can be found in Chapter \ref{Performance}. Algorithms that are specialised to solve Sudokus, for example by using an backtracking algorithm or stochastic approaches \cite{SolveSudoku2, SolveSudoku1}, perform significantly better than the case study implementation with the selection monad. It therefore shows that it is generally possible to solve puzzles like Sudoku with the selection monad; however, this approach is not feasible for solve Sudokus with more than 15 gaps.
Similar to the Connect Three implementation as described in Section \ref{ConnectThreeDiskussion}, the use of the selection monad, in combination with the library, enables a concise implementation in around 100 lines of code. Only the board and a function that checks the validity of the board have to be defined. 
\section{Simplified Chess}
Chess is played by millions of players worldwide. Many computer scientist are working since the 1950s on chess AIs in order to build better and better chess AIs. Many of these chess AIs are based on the minimax algorithm, and the selection monad should therefore be suitable to implement a chess AI. However, to limit the complexity of the game for this case study, only a simplified chess version was implemented, focusing on solving end-games and only supporting the following chess pieces: King, Queen, Rook, and Bishop. 
\subsection{Implementation}
As before, the Board, Move, and Outcome types need to be defined. The chessboard is represented using a matrix of chess pieces, where a chess piece is either Nothing or a Queen, Rook, King, or Bishop of a specific colour. The colour is either white or black. A move is a tuple of two positions on the board. The first element is the origin position and the second is the target position. A position is a tuple of Integers representing the x and y values.  The outcome type is (similar to the Connect Four implementation) a tuple of Integers where the first Integer is tracking if the game is a win, loss, or draw. The second Integer is counting the moves that need to be done to achieve this outcome. The definition of these data types is the following:
\begin{figure}[h!]
\begin{lstlisting}[language=Haskell, style=myHaskellStyleInline]
type R = (Int, Int)
type Position = (Int, Int)
type Move = (Position, Position)
type Board = Matrix ChessPice
data Colour = W | B
  deriving (Eq, Show)
data ChessPice = N | Queen Colour | King Colour | Rook Colour | Bishop Colour
  deriving (Eq, Show)
\end{lstlisting}
\caption{Chess data type definitions}
\end{figure}\\
As this simplified version of chess does not supporting a draw, the win function can be easily implemented by checking the whole board to see if there is still a king on the board for the given colour. The implementation of this win function looks like this:
\begin{figure}[h!]
\begin{lstlisting}[language=Haskell, style=myHaskellStyle]
wins :: Board -> Colour -> Bool
wins b c =  King (changeColour c) `notElem` toList b
\end{lstlisting}
\caption{Chess wins function}
\end{figure}\\
In the contrast to the simple \textit{wins} function, the function calculating all possible moves is essentially implementing all the chess rules. In order to do so, it iterates through the whole board and combines all possible moves of each piece with the colour of the current player. The implementation of this function can be seen in Figure \ref{PossibleMovesChess}. The implementations that calculate all possible moves for each chess piece are extensive and will be omitted because of the limited space. However, the full implementation can be found in the supporting material.\\
\begin{figure}[h!]
\begin{lstlisting}[language=Haskell, style=myHaskellStyleInline]
getPossibleMoves :: [Move] -> [Move]
getPossibleMoves m = concat [getMoves (x,y) b c | x <- [1..8], y <- [1..8], getMoves (x,y) b c /= [] ]
  where
    b = insertMoves startBoard m
    c = if (length m `mod` 2) == 1 then B else W

getMoves :: Position -> Board -> Colour -> [Move]
getMoves p b c1 = case b ! p of
                     N          -> []
                     (King c2)   -> if c1 == c2 then getKingMoves p else []
                     (Queen c2)  -> if c1 == c2 then getQueenMoves p b else []
                     (Rook c2)   -> if c1 == c2 then getRookMoves p b else []
                     (Bishop c2) -> if c1 == c2 then getBishopMoves p b else []
\end{lstlisting}
\caption{Functions that calculate all possible moves of the current player in chess. } 
\label{PossibleMovesChess}
\end{figure}\\
To calculate the outcome of the chess game, a list of moves will be applied to a previously defined start board. Therefore, an \textit{outcome} helper function is defined, calculating the outcome based on the current colour, a list of moves, the board onto which these moves need to be inserted, and a count of previous moves.
\begin{figure}[h!]
\begin{lstlisting}[language=Haskell, style=myHaskellStyle]
value :: Board -> Int
value b  | wins b W  = 1
         | wins b B  = -1
         | otherwise = 0
         
outcome :: Colour -> [Move] -> Board -> Int -> R
outcome _ [] b i       = (value b, i)
outcome c (m : ms) b i = let nb = insert b m in
                         if wins nb c then (value nb, i+1) else outcome (changeColour c) ms nb (i+1)

p :: [Move] -> R
p ms = outcome W ms startBoard 0
\end{lstlisting}
\caption{Chess outcome function}
\end{figure}\\
The \textit{eplsions} function is the same as in the Connect Three version, describing the minimax algorithm with alternating minimum and maximum function. It is limited to search only six moves in advance. This parameter needs to be adjusted to fit each endgame individually.
\begin{figure}[h!]
\begin{lstlisting}[language=Haskell, style=myHaskellStyle]
epsilons :: [[Move] -> J R Move]
epsilons = take 6 all
  where all = epsilonO : epsilonX : all
        epsilonX history = epsilonMinTupleParalell (getPossibleMoves history)
        epsilonO history = epsilonMaxTupleParalell (getPossibleMoves history)
\end{lstlisting}
\caption{Chess epsilons function}
\end{figure}
\subsection{Discussion}
The implementation is able to solve a simple endgame with a checkmate in six moves in under a minute. Every additional move that is computed in advance increases the computation time by a factor of about 14. A detailed discussion about the performance of this implementation can be found in Chapter \ref{Performance}.\\
The simplified version of chess shows that it is possible to implement chess with the help of the selection monad. The implementation of the simplified rules, as well as the AI, is done with around 140 lines of code, and is very concise. As the outcome function is only considering wins, draws, and losses, the implementation can only solve end games that lead to a win in six moves in a reasonable amount of time.\\
In general, this case study shows that even sophisticated games can be implemented to some extent with the selection monad. It also enables an elegant way of defining the behaviour of the AI. In order to expand this case study to a chess AI that is challenging to play against, all chess rules need to be implemented and the outcome function should be expanded to a function that takes the current positioning on the board into account.\\
When comparing the approach to other implementations of chess AI's, many implementations also use the minimax algorithm. To reduce the search space, the minimax algorithm is often combined with alpha/beta pruning, which could be a possible extension to the library and is discussed further in Chapter \ref{Discussion}. The key element for making a good chess AI is the outcome function. Having an efficient outcome function that is also taking the current positioning into account could significantly increase the difficulty of the AI. 

%%%%%%%%%%%%%%%%%%%%%%%%%%%%%%%%%%%%%%%%%%%%%%%%%%%%%%%%%%%%%%%%%%%%
% Einleitung
%%%%%%%%%%%%%%%%%%%%%%%%%%%%%%%%%%%%%%%%%%%%%%%%%%%%%%%%%%%%%%%%%%%%
% 6 Pages

\chapter{Performance and Testing}\label{Performance}
In order to measure the performance, each case study was tested with an example input and the computation time was measured for each layer of moves computed in advance. The computation time grows exponentially with each layer of moves added to the computation. Therefore, there will always be a limit after which it is not feasible to add another layer of moves. The only chance to achieve good results is, therefore, to push this limit as high as possible. As a result of the performance analysis, three major factors that influence the performance were identified. First, the amount of moves that are possible at each layer; second, the complexity of the functions that calculate the outcome and the possible moves; and third, an efficient minimum and maximum computation, for example using parallelism.\\
The following Chapter will go through each case study and explain the actions that were taken to increase the performance of the implementation in respect to the previously introduced factors influencing the performance. For each step of the optimisation, the performance was measured and is displayed in a graph. \\
All performance measurements were executed on the same machine with an AMD FX-8350 eight core processor with 4.00 GHz and 16 GB of RAM. The different versions with different optimisations can be found in the supporting material.\\
The final section of this chapter explains how the library was tested.
\section{Connect Four}
\begin{figure}[h!]
\centering
\includegraphics[width = .8\textwidth]{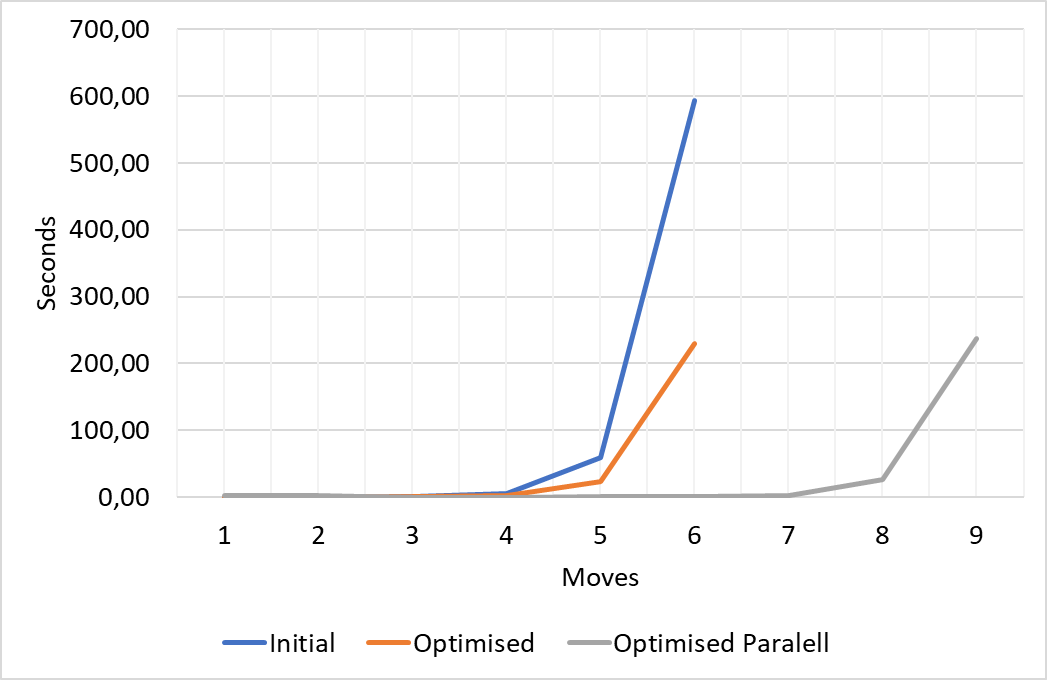}
\caption{Performance of the different Connect Four implementations}
\label{PerformanceConnectFour}
\end{figure}
\begin{figure}[h!]
\centering
\includegraphics[width = .8\textwidth]{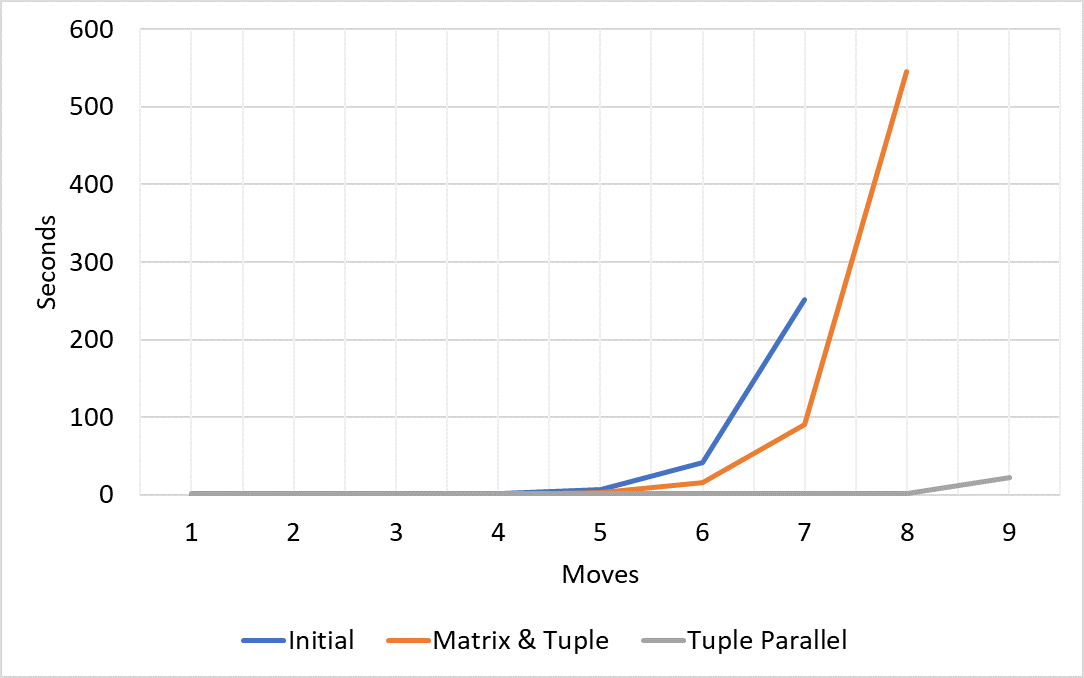}
\caption{Performance of the different Connect Three implementations}
\label{PerformanceConnectThree}
\end{figure}

A quick performance analysis of an initial implementation of Connect Four showed that, due to the exponentially growing computation time, it is not feasible to explore all possible moves. To reduce the search space, the simplified Connect Three version of the game was introduced. While an initial version of Connect Four was able to consider 5 moves in advance under 100 seconds, as seen in Figure \ref{PerformanceConnectFour}, the simplified Connect Three version was already able to consider 6 moves in advance in under 100 seconds, as seen in Figure \ref{PerformanceConnectThree}. The following performance optimisations where then applied to both versions: The Connect Three and Connect Four.
\subsection{Matrix instead of Lists}
In order to improve the performance, a matrix library was used instead of nested lists to represent the game board. This library enables access times of $O(1)$. This library, in combination with small adjustments to the function that calculates the outcome of the game, increased the performance slightly. The increased performance can be seen in Figure \ref{PerformanceConnectFour} and Figure \ref{PerformanceConnectThree}.
\subsection{Using parallel min max}
To increase the performance further, a parallel version of the minimum and maximum function was used, which led to a huge performance increase. The parallel version of Connect Four was able to consider 8 moves, instead of 5, in advance in under 100 seconds, and in the Connect Three version with 9 moves instead of 6. 
\subsection{Discussion}
With these performance optimisations, it was possible to compute a perfect play for the simplified Connect Three version in 22 seconds. This shows that an optimised outcome function and a reduction of the possible move for each layer improves the performance slightly. Using parallelism to compute the minima and maxima greatly benefits performance. Unfortunately, even for the optimised parallel Connect Four version, the estimated time of calculating a perfect play is about $2*10^{26}$ years and is still not feasible. However, the interactive version of Connect Four showed that it is possible to achieve good results when only considering 6 moves in advance. Therefore, the selection monad is suitable to build an interactive version of Connect Four and is even able to calculate a perfect play for the simplified Connect Three version.
\section{Sudoku}
\begin{figure}[h!]
\centering
\includegraphics[width = .8\textwidth]{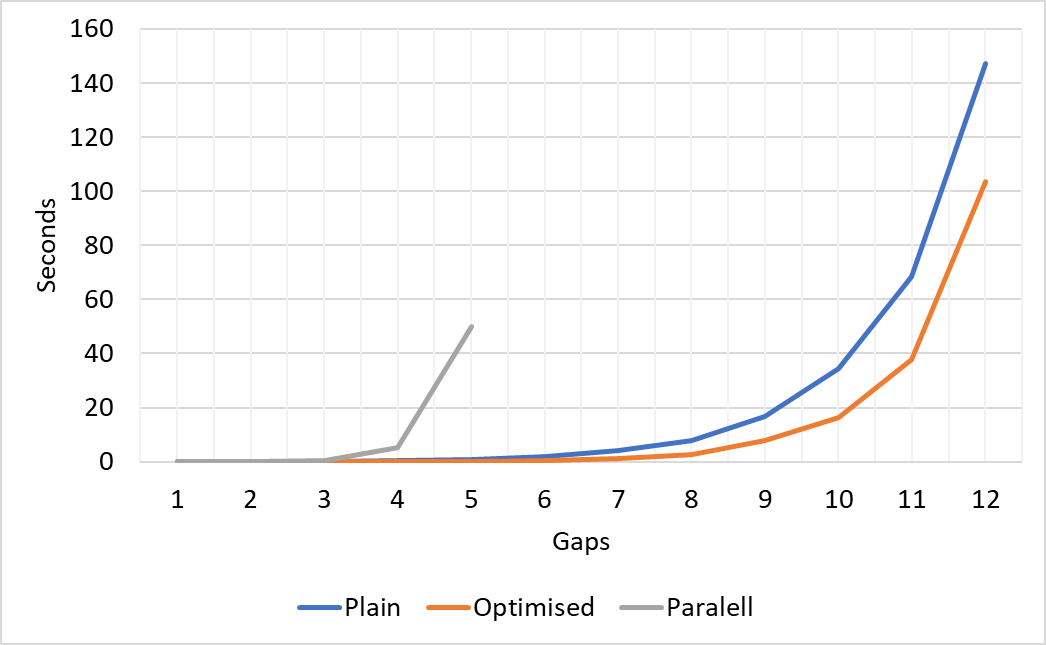}
\caption{Performance of the different Sudoku implementations}
\label{PerformanceSudoku}
\end{figure}

Solving Sudokus is known to be NP complete. However, there exist some heuristc and stochastic approaches \cite{SolveSudoku1, SolveSudoku2} that can solve Sudokus relatively efficient. Compared to those algorithms, the case study implementation is using a brute force approach that tries all possible combinations. This approach results in an exponential computation time. As shown in Figure \ref{PerformanceSudoku}, the unoptimised version can solve a Sudoku with 11 gaps in under 100 seconds where the optimised version can solve Sudokus with 12 gaps in under 100 seconds.
\subsection{Optimised version}
In order to optimise the computation, the function that calculates if a Sudoku is valid was optimised. Instead of checking the validity of the whole board after inserting a move into the board, only the row, column, and square in which a number is inserted is checked for validity.
\subsection{Parallel version}
In the case of Sudoku, the parallel version of the implementation performed significantly worse than the other implementations. A reason for this might be that the minimum and maximum functions, which work for boolean as outcome type, are only searching the list of moves until they find the correct move. In this case, the pruning used in the non-parallel versions of the boolean minimum and maximum function is more efficient than using an parallel version of the maximum function. Also, the parallel version is only using around 3\% to 5\% of the CPU where, in the other parallel versions of the case studies, the CPU was used up to 100\%.
\subsection{Discussion}
The Sudoku case study strengthens the hypothesis that an efficient outcome function has an impact on the performance. However, the simple brute force approach that the selection monad provides is not good enough to solve sophisticated Sudoku puzzles. It also shows that the right choice of minimum and maximum function can have a huge impact on the performance (see Figure \ref{PerformanceSudoku}). The functions that are optimised for Booleans work significantly better than the generic parallel version. As the selection monad only enables this brute force approach and does not allow for customisation of the AI's strategy to something that works well for Sudokus, the selection monad might not be suitable to solve sophisticated Sudoku puzzles.\\
However, this case study shows that it is generally possible to solve single player puzzles with a brute force approach, and the only limitation is the complexity of this puzzles. To resolve this performance issue, the selection monad could be modified to support some sort of pruning (like alpha-beta pruning). As this pruning optimisation is not in the scope of this thesis, it will briefly discussed in the Future Work Section \ref{FutureWork}.
\section{Chess}
The building of chess AIs has always been a challenge in computer science. The complexity of the game is huge, and experts are not sure whether it can be solved, or if a perfect play can be found. Because of the complexity, only a simplified version of chess has been implemented in this case study. This simplified version has an easy outcome function that is essentially only checking if both kings are still on the board. However, the huge amount of possible moves affects the performance of this implementation the most. Of the different performance optimisations already studied in detail in the other two case studies, only one implementation of the chess game exists. The aim of this case study is less to explore the performance problems of the selection monad implementations, and more to show that even sophisticated games like chess can be implemented to some extent with the help of the selection monad. 
\subsection{Runtime}
\begin{figure}
\centering
\includegraphics[width = .8\textwidth]{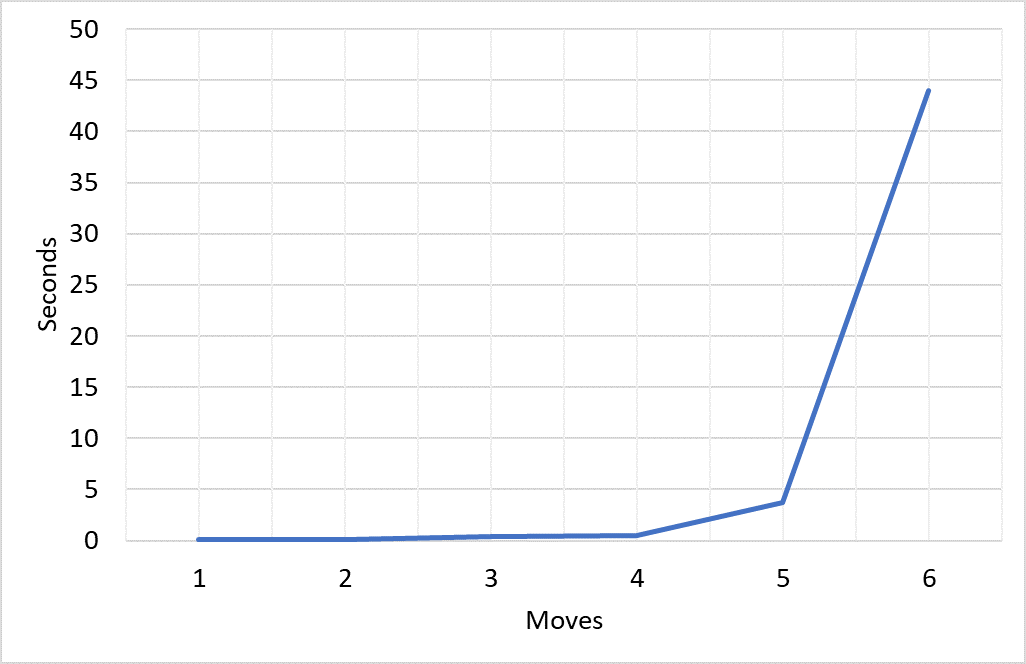}
\caption{Performance of the simplified Chess implementations}
\label{PerformanceChess}
\end{figure}

The performance of the Chess implementation, as seen in Figure \ref{PerformanceChess}, shows that the chess AI can only calculate up to six moves in advance within reasonable time. As the outcome function is already as simple as possible, this performance is due to the huge amount of moves that add to the computation with each layer.
\section{Discussion}
Again, the selection monad needs to be extended to support some sort of pruning (like Alpha-Beta pruning). This is discussed, together with the Sudoku implantation improvements, as mentioned in Section \ref{FutureWork}. However, this implementation is already able to solve simple end-games, considering 6 moves in advance in under 100 seconds. Considering 7 moves in advance increases the computation time to over 1000 seconds. With a more sophisticated outcome function, this AI can easily be extended to an interactive AI that calculates up to 6 moves in advance. This shows that the selection monad is already able to support an implementation of a chess AI and, with some pruning extensions to the selection monad, the number of moves that are calculated in advance might increase further.
\section{Testing of the library}
The selection monad library is tested with unit tests using the HUnit test library \cite{HUnit}. These Unit tests cover the different minimum and maximum functions, as well as some helper functions regarding the selection monad. The functions that calculate the optimal play need an actual game implementation and are therefore covered and tested with the case studies. An explanation for how to execute the tests, as well as the source code of the tests, can be found in the supporting materials.

%%%%%%%%%%%%%%%%%%%%%%%%%%%%%%%%%%%%%%%%%%%%%%%%%%%%%%%%%%%%%%%%%%%%
% Diskussion und Ausblick
%%%%%%%%%%%%%%%%%%%%%%%%%%%%%%%%%%%%%%%%%%%%%%%%%%%%%%%%%%%%%%%%%%%%
% 10 Pages

\doublespacing
\chapter{Discussion and Outlook}\label{Discussion}
This Chapter briefly summarises the work done for this the masters project in Section \ref{Summary} and discusses the major performance issue that occurred during the implementation of the case studies in Section \ref{Performance}. The results of the case studies are summed up in Section \ref{Conclusion}, while Section \ref{FutureWork} concludes this thesis with an short introduction of further work that could be done.
\section{Summary}\label{Summary}
At the beginning of the project, literature research was done regarding the selection monad. During this phase, a deep understanding of the mechanics of the selection monad was gained. With this knowledge, the three case study games were implemented. In order to increase the performance of the case studies, optimisations were applied to the outcome functions of the case studies. Simultaneously, the library was built, summarising common functionalities of the case studies. Additionally, performance optimisations to the minimum and maximum functions were implemented in the library. During this implementation phase, general knowledge in game theory was gained, especially regarding the implementation of game AIs. Furthermore, expertise in library design was gained during the library building process.\\
Finally, a performance benchmark was done, investigating the case studies with their different optimisations. This was done in order to identify the key factors influencing the implementation the most.  
\section{Performance}\label{Performance}
The complexity of the games was identified as a major factor when deciding whether the selection monad is a suitable approach or not. The minimax algorithm used as the underlying algorithm of the selection monad has an exponential computation time. Therefore, currently the only way to increase the performance is to provide a fast implementation of the outcome function and the use of the right minimum and maximum functions, which ideally allow for parallelisation of the computation.
\section{Conclusion}\label{Conclusion}
The selection monad provides an elegant way of implementing sequential games and puzzles. The case studies show that the implementation of a sequential game can be done in under 150 lines of code. Furthermore, with the library a programmer does not need to care about how to implement the AI. The AI can be implemented just by defining an outcome function and using the provided minimum and maximum functions. While implementing sequential games, the complexity of the games was identified as the only issue, resulting in performance problems. This issue could potentially be resolvable by including a pruning mechanism into the library.\\
Summarising common functionalities into a library improves the modularity of case studies code significantly, and future programmers profit from the performance optimisations that are built into the minimum and maximum functions. They, therefore, only need to choose the right functions that fit their outcome type.
\section{Future work}\label{FutureWork}
To extend the functionality of the library, a possible future work could be implementing more case studies to explore a more diverse set of games. In this thesis, only games with perfect knowledge (i.e. games where every player is perfectly informed about all events and moves that where done previously) where considered. Therefore a possible future work could explore games without perfect knowledge, where some part of the game is hidden to the player. This, for example, could be a card game or Mastermind. Furthermore, a possible future case study could explore the ability of the selection monad to play games that are based on chance. An example for this kind of games could be Black Jack.\\
Additionally,  future work could explore performance optimisations that can be applied to the library. For example, pruning algorithms could limit the search space of moves that need to be explored. Alpha-beta pruning, for example, is an pruning algorithm that skips whole branches of the tree during the minimax algorithm. It skips the evaluation of a branch in the tree when at least one possible move has been found that shows that this branch is worse than the previous explored moves. Adding pruning algorithms to the library might increase the performance further, making the selection monad a useful tool to implement AIs for sequential games.
% TODO: Add ending in here

\singlespacing
%%%%%%%%%%%%%%%%%%%%%%%%%%%%%%%%%%%%%%%%%%%%%%%%%%%%%%%%%%%%%%%%%%%%%%%%%%%%%
%%% Appendix
%%%%%%%%%%%%%%%%%%%%%%%%%%%%%%%%%%%%%%%%%%%%%%%%%%%%%%%%%%%%%%%%%%%%%%%%%%%%%
\appendix

%\setcounter{secnumdepth}{-1}
%\section{Tables}\label{chap:App}

%%%%%%%%%%%%%%%%%%%%%%%%%%%%%%%%%%%%%%%%%%%%%%%%%%%%%%%%%%%%%%%%%%%%%%%%%%%%%
%%% Bibliographie
%%%%%%%%%%%%%%%%%%%%%%%%%%%%%%%%%%%%%%%%%%%%%%%%%%%%%%%%%%%%%%%%%%%%%%%%%%%%%

\addcontentsline{toc}{chapter}{Bibliography}

\bibliography{mylit}
\bibliographystyle{unsrt}
%% Obige Anweisung legt fest, dass BibTeX-Datei `mylit.bib' verwendet
%% wird. Hier koennen mehrere Dateinamen mit Kommata getrennt aufgelistet
%% werden.

\end{document}